\documentclass{article}

\PassOptionsToPackage{numbers,sort&compress}{natbib}

\usepackage[preprint]{neurips_2025}

\usepackage[utf8]{inputenc} 
\usepackage[T1]{fontenc}    
\usepackage{hyperref}       
\usepackage{url}            
\usepackage{booktabs}       
\usepackage{amsfonts}       
\usepackage{nicefrac}       
\usepackage{microtype}      
\usepackage{xcolor}         
\usepackage{makecell}
\usepackage{amsmath}
\usepackage{amsthm}
\usepackage{amssymb}
\usepackage{graphicx}
\usepackage{subcaption}  
\usepackage{multirow}
\usepackage{colortbl}
\usepackage{spverbatim}

\newcommand{\passk}{\textrm{Pass@}k}
\newcommand{\E}[1]{E \! \left\{ #1 \right\}}

\newcommand{\Repeater}{\textrm{Repeater}}
\newcommand{\Variator}{\textrm{Variator}}

\newtheorem{theorem}{Theorem}

\title{Leveraging LLM Inconsistency \\ to Boost Pass@$k$ Performance}

\author{
  Uri Dalal
  \quad
  Meirav Segal 
  \quad
  Zvika Ben-Haim
  \quad
  Dan Lahav
  \quad
  Omer Nevo \\
  Pattern Labs \\
  \texttt{\{uri,meirav,zvika,dan,omer\}@patternlabs.co}
}

\begin{document}

\maketitle

\begin{abstract}
Large language models (LLMs) achieve impressive abilities in numerous domains, but exhibit inconsistent performance in response to minor input changes. Rather than view this as a drawback, in this paper we introduce a novel method for leveraging models’ inconsistency to boost $\passk$ performance. Specifically, we present a ``Variator'' agent that generates $k$ variants of a given task and submits one candidate solution for each one. Our variant generation approach is applicable to a wide range of domains as it is task agnostic and compatible with free-form inputs. We demonstrate the efficacy of our agent theoretically using a probabilistic model of the inconsistency effect, and show empirically that it outperforms the baseline on the APPS dataset. Furthermore, we establish that inconsistency persists even in frontier reasoning models across coding and cybersecurity domains, suggesting our method is likely to remain relevant for future model generations.
\end{abstract}

\section{Introduction}
\label{sec:intro}
The capabilities of large language models (LLMs) have grown tremendously in the past few years, enabling their application across diverse domains, from marketing and sales to software engineering~\cite{mckinsey2023generativeai}. Given this progress, it is perhaps surprising that LLMs exhibit specific forms of reasoning limitations. In particular, in this paper we focus on the \emph{inconsistency effect,} whereby a model's success at a particular task can be strongly affected by semantically equivalent variations in the input prompt, such as replacing a word with a synonym~\cite{leidinger2023language, gu2022robustness}.
This inconsistency has been observed to significantly affect performance in a wide range of domains~\cite{pezeshkpour2023large, alzahrani2024benchmarks, li2024perteval, zhuo2024prosa, voronov2024mind}.

Due to reliability requirements in practical tasks, inconsistency in language models is generally viewed as undesirable, with studies mostly focusing on mitigating the effect~\cite{mizrahi2024state, honovich2022unnatural, sun2023evaluating, sclar2024quantifying, alzahrani2024benchmarks}. By contrast, we show that this seemingly problematic effect can be used advantageously to improve model performance, as measured by the $\passk$ metric~\cite{chen2021code}. This metric measures performance when the model is allowed to submit $k$ solution candidates, with success counted if at least one such solution is correct~\cite{chen2021code}. Such a scenario makes sense when solutions can readily be tested for correctness, as occurs in many coding tasks (which can be evaluated on a test suite~\cite{chen2021code, kulal2019spoc}), as well as in some cybersecurity tasks (e.g., fuzzing~\cite{liu2023aipowered}).

\begin{figure}
    \centering
    \begin{subfigure}[b]{0.41\textwidth}
        \centering
        \includegraphics[width=1.75in]{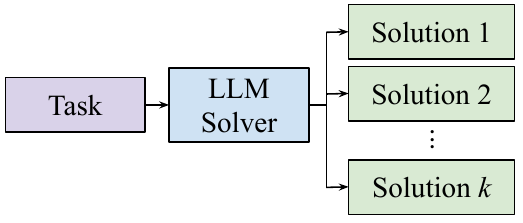}
        \caption{The baseline Repeater agent for a $\passk$ task generates $k$ candidate solutions by sampling $k$ responses from the LLM\@.}
        \label{fig:overview-repeater}
    \end{subfigure}
    \hspace{1em}
    \begin{subfigure}[b]{0.55\textwidth}
        \centering
        \includegraphics[width=2.8in]{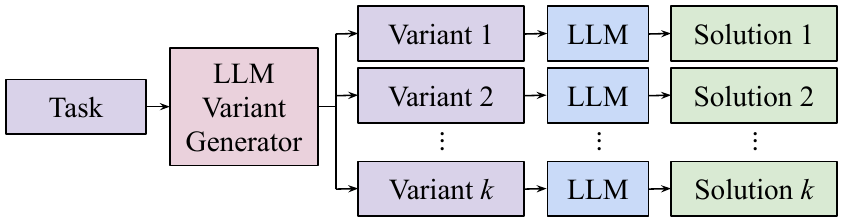}
        \caption{We propose instead to use the Variator agent, which first builds $k$ equivalent variants of the original task, then generates a single candidate solution for each variant.}
        \label{fig:overview-variator}
    \end{subfigure}
    \caption{Overview of the proposed technique, which is shown both empirically (for a commonly used benchmark) and theoretically (under appropriate symmetry conditions) to improve the $\passk$ success rate relative to the standard approach.}
\label{fig:overview}
\end{figure}

We introduce a novel ``Variator'' agent (Figure~\ref{fig:overview-variator}), which generates $k$ variants of the original problem and submits a candidate solution to each variant. The variants are generated through a task-agnostic method whereby an LLM paraphrases the prompt and alters its theme. We theoretically demonstrate the benefit in performance of the Variator agent compared to a baseline (Figure~\ref{fig:overview-repeater}) in which $k$ solutions are generated for the original task (Section~\ref{sec:theoretical}). Critically, our analysis shows that improvement occurs even if the model is, on average, no more likely to solve individual variants than it is to solve the original challenge. Intuitively, this happens because the occasional improvements in performance achieved by some variants are amplified by the $\passk$ metric.  We also evaluate the Variator agent on a public benchmark, where model memorization is likely to hinder our method by artificially inflating the performance of the original challenge relative to its variants. We show that Variator's improvement is strong enough to overcome these headwinds and outperform the baseline (Section~\ref{sec:empirical}). On a private dataset created from variants of the original tasks, we show that Variator significantly outperforms the baseline.

With the release of more advanced models, a natural question arises regarding the persistence of the inconsistency effect and, consequently, the continued relevance of our proposed method. To the best of our knowledge, the inconsistency effect has not previously been demonstrated in reasoning models, which are trained to output intermediate steps prior to the task solution~\cite{jaech2024openai}; indeed, one might suppose that the long sequence of reasoning tokens between the input (task definition) and the output (the proposed solution) would result in reduced sensitivity to minor changes in prompt formulation. Nevertheless, we demonstrate that the effect persists even in frontier reasoning models, such as OpenAI's o3-mini~\cite{openai2025o3mini} and Claude 3.7 Sonnet (in extended thinking mode)~\cite{anthropic2025sonnet37}. We focus on two domains, coding and cybersecurity, which were chosen because they require complex, long-form solutions, yet can be automatically tested. This result adds to existing findings that larger models are not necessarily less sensitive to minor task changes~\cite{sclar2024quantifying, lu2021fantastically, leidinger2023language, chatterjee2024posix, zhuo2024prosa}, and therefore the inconsistency effect may persist in future models as well.

Our contributions in this paper are as follows: (1) we provide a novel Variator agent, which relies on an automated method for generating variants of the original task; (2) we prove analytically that Variator improves models' $\passk$ performance, under plausible assumptions on the inconsistency effect; (3) we validate our approach empirically on a public dataset, achieving performance improvement despite the counteracting influence of memorization effects, and on a private dataset, significantly outperforming the baseline; and (4) we demonstrate that LLM inconsistency, previously shown for simpler tasks and non-reasoning models, continues to hold for complex, free-form challenges and with frontier reasoning models. This work offers a novel perspective on how we might leverage rather than merely mitigate LLM inconsistency.

\section{Related work}
\label{sec:related}

The inconsistency effect in LLMs, also referred to as brittleness~\cite{mizrahi2024state} or prompt sensitivity~\cite{habba2025dove}, has been demonstrated in several studies over the past few years, across multiple models, domains, and variation types~\cite{leidinger2023language, gu2022robustness, pezeshkpour2023large, alzahrani2024benchmarks, li2024perteval, zhuo2024prosa, voronov2024mind}.  In most studies, variations are generated for closed-form tasks such as multiple-choice questions, likely due to the ease of answer verification. ~\citet{chatterjee2024posix} examined the inconsistency effect in open-ended questions, but only by quantifying the difference between outputs, regardless of any performance score. Meanwhile, ~\citet{zhuo2024prosa} reported that LLMs exhibit greater performance inconsistency for tasks that require long output, such as coding tasks. In our work, we examine inconsistencies for coding challenges as well as for the previously unexamined domain of cybersecurity.

The task variations presented in prior studies can be categorized into three groups: structural changes (such as adding white\-space~\cite{voronov2024mind} or order shuffling of few-shot examples~\cite{sclar2024quantifying, lu2021fantastically, voronov2024mind}), grammatical properties (e.g., tense and synonym replacements~\cite{leidinger2023language, gu2022robustness}), and paraphrasing (a free-form change that preserves meaning~\cite{gu2022robustness, gonen2022demystifying, chatterjee2024posix}). The use of free-form inputs (e.g., a coding task specification) allows us to consider variations in a wider sense, which is agnostic to the challenge domain (Section~\ref{sec:generating}). We thus extend the paraphrasing category to include wording, terminology and backstory modifications.

Other than pointing out inconsistency as a problem, some studies attempted to find the best-performing prompt~\cite{habba2025dove, gonen2022demystifying}. However, such methods have limited generalization because specific prompt variations may increase the success rate of one model while reducing it for another~\cite{habba2025dove, voronov2024mind, alzahrani2024benchmarks, sclar2024quantifying, lu2021fantastically, leidinger2023language}. The concept of leveraging inconsistency to enhance model performance was previously examined by \citet{voronov2024mind}, who suggested generating predictions across multiple variants, and outputting the label with the highest average probability. However, this voting approach is relevant only for classification tasks, and assumes that the most common response is the correct one. By contrast, our setting is appropriate for any solution format, including long-form output, and as we shall see, the $\passk$ metric improves even if the model only rarely identifies the correct solution.

\section{Experimental setup}
\label{sec:experimental_setup}

We illustrate the inconsistency effect when LLMs are tasked with solving variants of coding exercises and of cybersecurity capture-the-flag (CTF) competitions, and examine performance of OpenAI o3-mini~\cite{openai2025o3mini} and Claude 3.7 Sonnet (in extended thinking mode)~\cite{anthropic2025sonnet37}; the complete model settings are described in Appendix~\ref{app:compute}. In this section, we provide details of the setup used for these demonstrations.

\subsection{Challenges}
\label{sec:challenges}
We focus on two domains: coding and cybersecurity. For both domains, we consider open-ended questions, which require relatively long solutions. These domains were chosen due to the ability to automate the verification of challenge solutions, which reduces both costs and biases relative to manual verification. Both domains are frequently studied in the literature when evaluating LLM capabilities~\cite{hendrycks21apps, shao24nyuctf}.

For our coding experiments, we select coding tasks from the APPS benchmark~\cite{hendrycks21apps}. This dataset comprises 10,000 programming problems, stratified across three difficulty levels (introductory, interview, and competition). Each challenge includes a test suite for automatic solution verification.

For our cybersecurity demonstrations, we use CTF challenges, a method commonly used for evaluating LLM cybersecurity capabilities~\cite{anthropic2025sonnet37, openai2025o3mini, openai2025o3}. In these challenges, participants must recover hidden information (known as a ``flag'') through interactions with a vulnerable machine in a sandboxed environment. Obtaining the flag constitutes proof of successful challenge completion, allowing for automatic success verification. Unlike coding challenges, in which the LLM generates the full solution in a single output, CTF tasks typically require a multi-step interaction. In each step, the LLM engages with the environment through an agent, which allows the LLM to run commands and code on a ``local'' machine from which access is provided to the vulnerable target server. The LLM is allowed to install and use standard tools and packages on the local machine. As is common practice~\cite{rodriguez2025framework}, we limit the number of interactions and the time span of each interaction. We demonstrate the inconsistency effect on two novel CTF challenges described in Appendix~\ref{app:cyber_challenges}.

\subsection{Variant generation}
\label{sec:generating}

We refer to a variant as \emph{equivalent} if any solution to the variant is also a solution to the original task and vice versa, and the two are nearly identical in difficulty, as judged by a subject matter expert. Otherwise, the variant is said to be non-equivalent. This definition includes a wider range of variation than previously considered when examining the inconsistency effect. For example, in Table~\ref{table:coding-variant} of Appendix~\ref{app:variant_generation} we showcase two variants of a coding problem. These variants differ in backstory (changing from tiling a royal causeway to painting an interstellar vessel), wording, and variable names ($p,q$ vs. $a,b$). Nonetheless, careful reading of the variants shows that the underlying task is identical in both variants, making these variants equivalent.

We generate variants automatically using an LLM. For each experiment, the same model is used to generate the variants and to solve them; this is particularly important for Section~\ref{sec:leveraging}, where variants are generated and then solved as part of the proposed Variator agent. We employ a structured prompt that instructs the model to generate significantly altered versions of the original challenge (see details in Appendix~\ref{app:variant_generation}). For coding tasks, the prompt directs the model to modify multiple aspects of the challenge, including problem description, background story, language, and mathematical notation, while preserving the core problem structure. Critically, we constrain the model to maintain the original input-output format specifications to ensure functional equivalence between variants, so that the test suite for the original challenge would be applicable to the variants.

For the CTF challenge, we instruct the LLM to rewrite the vulnerable target application to have a different purported use, while retaining the same underlying functionality. We further ask the LLM to use a different software design for each variant, by maintaining an LLM-generated list of designs used for prior variants.  The original CTF challenges were designed by cybersecurity experts who also provided reference solution scripts that retrieve the correct flag. When used to verify the prevalence of the inconsistency effect, our prompt includes the solution to the original task, and directs the model to generate a variant that is solved using the same solution. This equivalence is automatically verified, so that if the original solution does not solve a proposed variant, then that variant is discarded.

While our approach successfully generates diverse variants, it does not inherently guarantee equivalence. For example, generated variants could potentially include hints rendering them easier to solve. It is also possible that variants might omit a crucial detail required for the variant to be solvable. To measure the likelihood that a generated variant is equivalent, we enlisted a cadre of professional cybersecurity experts, who conducted manual verification of generated variants. These experts were asked to highlight any aspects that might cause a variant to be perceived as either harder or easier than the original (see Appendix~\ref{app:manual_verification}). Variants failing one of these criteria represent 6\% of the variants for coding challenges and none of the variants for cyber challenges. Further details on this process are provided in Appendix~\ref{app:inconsistency-experiments}.

\section{The inconsistency effect}
\label{sec:inconsistency-effect}

In this section, we illustrate the inconsistency effect in reasoning models solving coding and cyber challenges by comparing a model's success rate on a challenge and on its variants. To this end, we focus here on a small number of challenges and their variants; later, in Section~\ref{sec:empirical}, we will further demonstrate the effect on a larger set of coding challenges, comprising all challenges from the APPS dataset with a sufficient number of unit tests.

\begin{figure}
    \centering
    \begin{tabular}{cc}
    \begin{subfigure}[b]{0.43\textwidth}
        \centering
        \includegraphics[width=\textwidth]{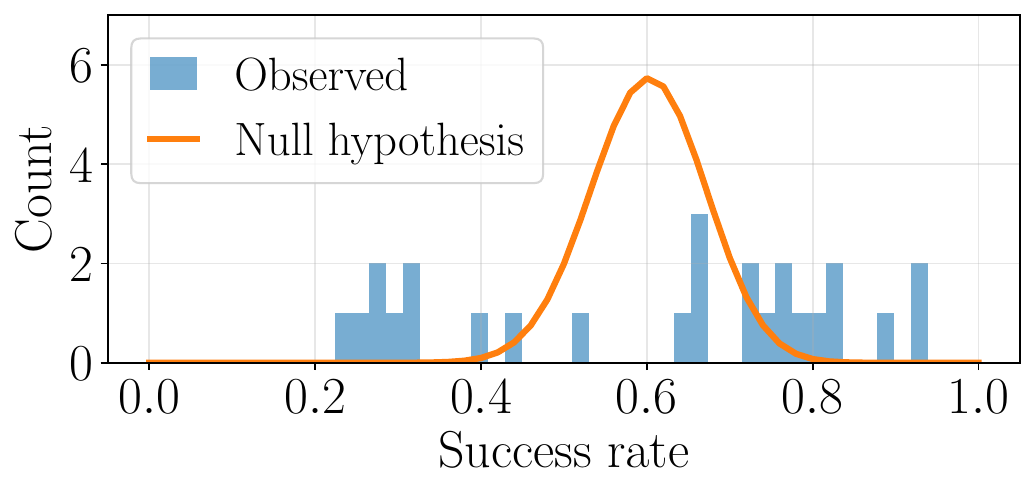}
        \caption{Coding experiment, Claude 3.7 Sonnet \\ (extended thinking mode). $p\textrm{-value} = 3 \times 10^{-6}$.}
    \label{fig:standalone_exp_claude_coding}
    \end{subfigure}
    &
        \begin{subfigure}[b]{0.43\textwidth}
        \centering
        \includegraphics[width=\textwidth]{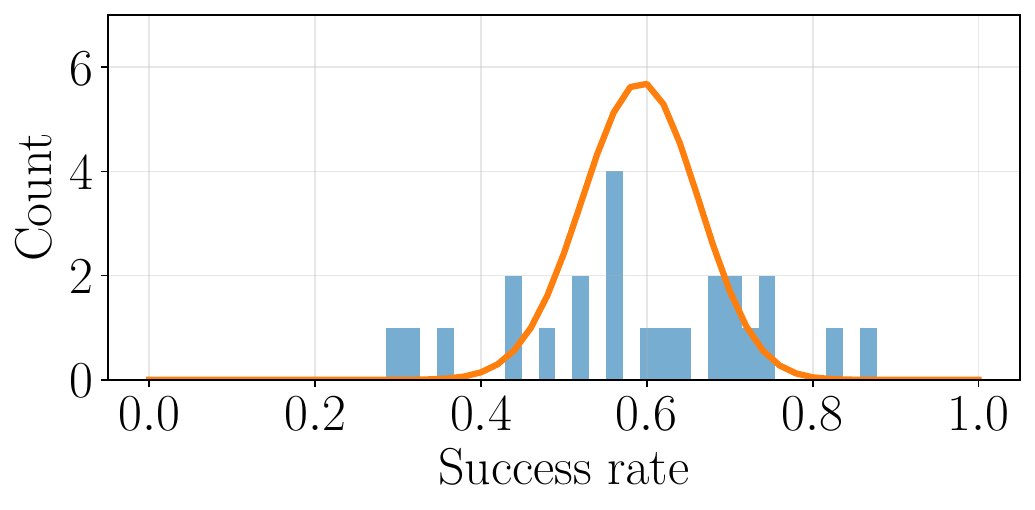}
        \caption{Coding experiment, OpenAI o3-mini. \\ $p\textrm{-value} = 4 \times 10^{-4}$.}
    \end{subfigure}
    \\[1em]
        \begin{subfigure}[b]{0.43\textwidth}
        \centering
        \includegraphics[width=\textwidth]{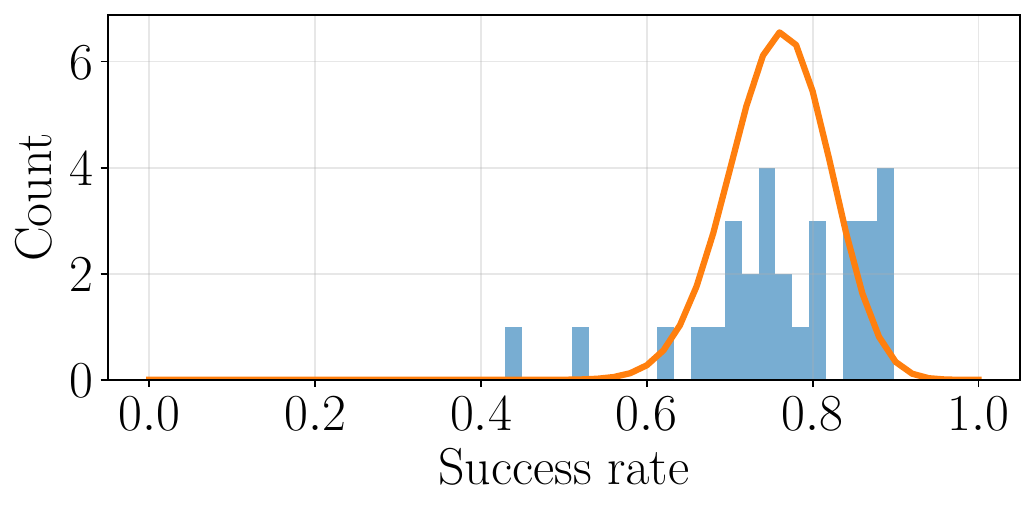}
        \caption{Cyber experiment, Claude 3.7 Sonnet \\ (extended thinking mode). $p\textrm{-value} = 5 \times 10^{-5}$.}
    \end{subfigure}
    &
        \begin{subfigure}[b]{0.43\textwidth}
        \centering
        \includegraphics[width=\textwidth]{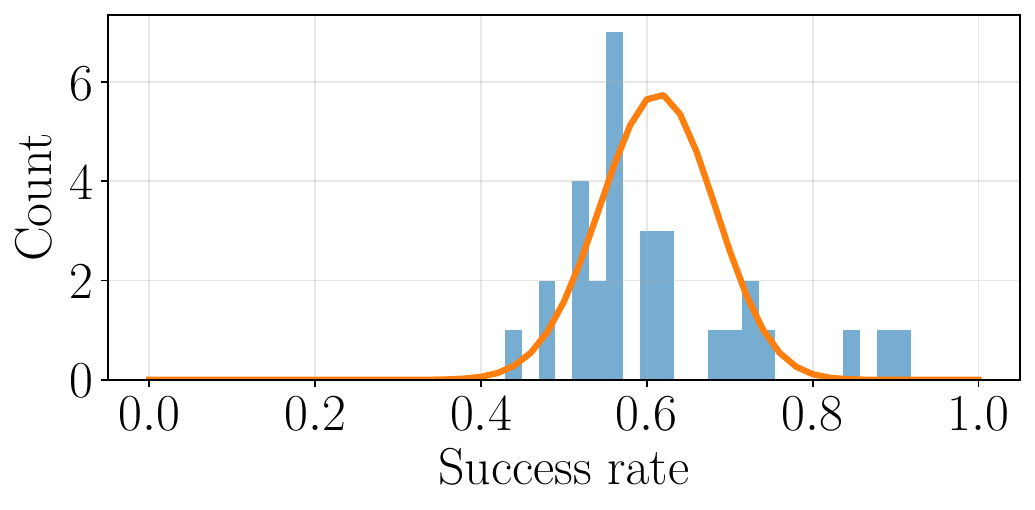}
        \caption{Cyber experiment, OpenAI o3-mini. \\ $p\textrm{-value} = 8 \times 10^{-5}$.}
    \end{subfigure}
    \end{tabular}
    \caption{Variant success rate distributions. Blue bars represent the observed histogram of variant success rates. Orange lines represent the expected distribution under the null hypothesis that there is no inconsistency effect in the given setting.}
\label{fig:standalone_exp}
\end{figure}

For each domain (cyber or coding) and for each model, we choose a single ``original'' challenge whose success rate was neither 0 nor 1 (see Appendices~\ref{app:cyber_challenges} and~\ref{app:inconsistency-experiments}), and generate a set of 30 equivalent variants as described in Section~\ref{sec:generating}. We exclude from this set variants which were identified by experts as non-equivalent (Section~\ref{sec:generating}). For each variant, we then repeatedly prompt the LLM to generate 50 solution candidates, and compute the mean success rate for each variant. The histograms of these success rates are plotted in blue bars in Figure~\ref{fig:standalone_exp}.

Due to the randomness of LLM responses, it is possible that different variants will produce different success rates purely by chance, even without the inconsistency effect; indeed, under the null hypothesis of equal success rates for all variants, the observed number of successes would follow a binomial distribution $\mathrm{Bin}(n,p)$, where $p$ is the (hypothesized, unique) task success rate and $n$ is the number of repetitions per variant. The distribution of expected success rates is plotted as an orange line in Figure~\ref{fig:standalone_exp}. However, the observed success rates differ significantly from this expected distribution. In particular, we can identify variants for which the success rate would be highly improbable under the null hypothesis. To formally reject this null hypothesis, we employed a Monte Carlo method to estimate the $p$-value. The resulting $p$-values are below $5 \times 10^{-4}$ in all cases, indicating statistical significance of the inconsistency effect in the examined reasoning models
(see Table~\ref{tab:standalone_experiments} in Appendix~\ref{app:inconsistency-experiments}).

\begin{figure}
    \centering
    \includegraphics[width=0.5\textwidth]{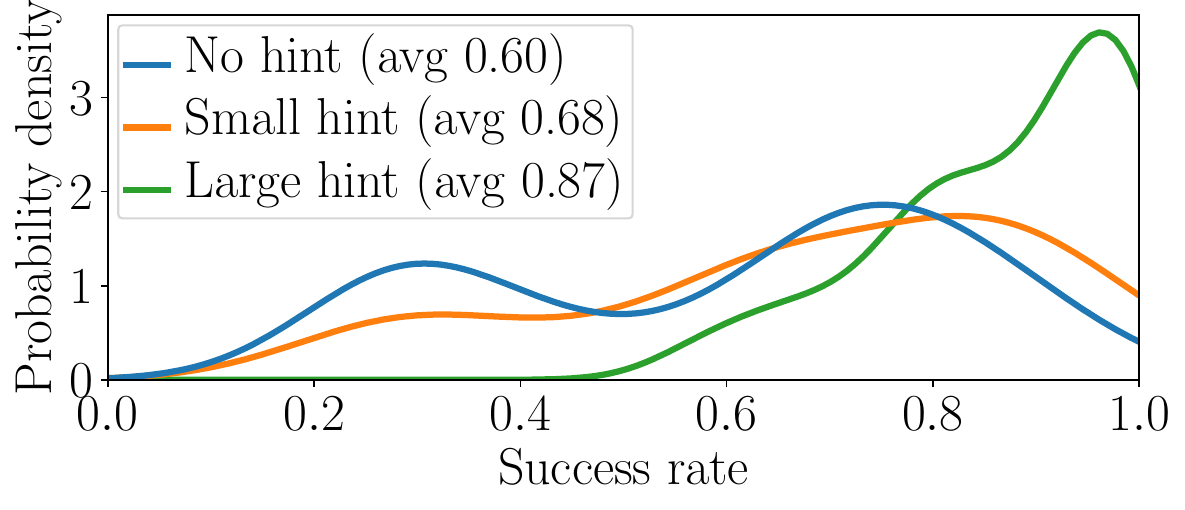}
    \caption{Distributions of success rates for 25 variants of the same challenge, with three guidance levels. The considerable overlap between the three distributions illustrates that some unguided variants have higher success rates than other guided variants, highlighting the significant performance impact of equivalent variants.}
\label{fig:hint_experiment}
\end{figure}

To provide an intuitive sense for the size of the inconsistency effect, we also examine the impact of explicit guidance added to the challenge: We create a set of 25 variants of a coding challenge, and evaluate model performance on these variants under three settings: (a) the variants without assistance; (b) each variant supplemented with a simple, single sentence hint; and (c) each variant supplemented with a detailed step-by-step guide to solving the challenge (see Appendix \ref{app:hint_experiment}). The average success rates of these settings were 60\%, 68\%, and 87\%, respectively. Figure \ref{fig:hint_experiment} illustrates the significant overlap between performance distributions across these settings: The intersection-over-union (IoU) between the unguided and small hint distributions is 82\%, while the IoU between the unguided and large hint distributions is 45\%. This indicates that some unguided variants have higher success rates than other guided variants, highlighting the substantial impact that equivalent variants can have on model performance, even relative to deliberately crafted guidance.

\section{Leveraging inconsistency to enhance LLM performance}
\label{sec:leveraging}
Thus far, we have established that model performance fluctuates considerably when given equivalent variants as input, even for frontier reasoning models. In many cases, this can rightly be seen as a limitation of LLMs, since one would want a model's ability to solve challenges to be independent of factors such as the background story in which the challenge is presented. 

However, this effect can also be advantageous, as we now show by presenting a novel agent that seeks to utilize the model's fluctuating success rate under the $\passk$ metric. As noted in Section~\ref{sec:intro}, this metric tests $k$ candidate solutions from the model and declares success if at least one is correct. It is often used when proposed solutions can be readily checked for correctness, such as for coding benchmarks~\cite{chen2021code} and cyber benchmarks~\cite{shao24nyuctf}.

To boost a model's $\passk$ performance, we propose a ``Variator'' agent, which operates as follows. Given an input (``original'') challenge, we generate $k$ variants (as described in Section~\ref{sec:generating}). As this technique is intended to be used as an autonomous agent, no human verification is performed on these variants. We then generate and submit one candidate solution from each variant (see Figure~\ref{fig:overview}). Assuming the generated variants are equivalent, a solution to any variant will also solve the original challenge. We compare Variator with a baseline ``Repeater'' agent, which generates and submits $k$ solution candidates from the original challenge.

Due to the inconsistency effect, some of the generated variants can potentially have a higher success rate than the original challenge, while others might have a lower success rate. At first glance, this would lead us to expect that Variator will have the same average performance as Repeater. This will indeed be the case if we merely generate a single variant and thence a solution (i.e., in the case $k=1$). However, we demonstrate both analytically and empirically that this is not the case for $k>1$. Theoretically, we show that under a simplified (but symmetric) model of variant performance, Variator achieves high performance on all challenges, while only slightly underperforming on easy challenges (Section~\ref{sec:theoretical}). Empirically, we demonstrate that $\passk$ performance of both Claude 3.7 Sonnet (extended thinking mode) and o3-mini improves on a subset of the APPS coding benchmark dataset~\cite{hendrycks21apps} when using Variator, compared with the Repeater baseline (Section~\ref{sec:empirical}).

\subsection{Theoretical analysis}
\label{sec:theoretical}

Suppose we are given a challenge $C$, and consider an LLM which can produce candidate solutions for $C$. Denote by $p_o$ the success rate of such a randomly sampled candidate. We wish to analyze the performance of the Repeater and Variator agents described above.

The $\passk$ performance of these two agents depends on the difficulty of the original challenge $C$ and its variants. For example, if the variants tend to be easier to solve than the original challenge, then we would not be surprised to find that Variator outperforms Repeater. Surprisingly, though, even under perfectly symmetrical conditions, Variator may still have a significant advantage over Repeater. 

An intuitive explanation for this is as follows: For ``easy'' challenges, where $p_o$ is close to $1$, there will be some variants which are rather harder (having lower success rate). But there cannot be challenges which are much easier, since the success rate is already close to $1$. Thus, variants of easy challenges may have a mean success rate which is somewhat lower than $p_o$. Conversely, variants of hard challenges may have a mean success rate which is somewhat higher than $p_o$. These two effects influence the $\passk$ metric differently: The fact that a single correct solution out of $k$ is considered a successful submission benefits small increases at low probabilities much more than it penalizes small decreases at high probabilities.

To show this formally, one must precisely define the probability distribution of success rates of a variant of $C$. We provide the theorem below as an example illustrating the aforementioned symmetry-breaking effect; analogous results can be shown for other settings.

\begin{theorem} \label{thm:guar}
Let $C$ be a challenge with success rate $p_o$, and consider a variant-generation mechanism which yields variants whose success rate is a random variable $P_v = [p_o + W]_0^1$, where W is uniformly distributed in the range $[-w, w]$ and $[\cdot]_0^1$ represents clipping to the range $[0,1]$. We refer to the constant $w$ as the \emph{spread} of the variants and assume for simplicity $0 < w < \tfrac12$. We then have, regardless of the value of $p_o$,
\begin{align}
\text{Performance guarantee:} \quad
\passk(\Variator) &\ge 1 - (1 - \tfrac{w}{4})^k, \label{eq:perf-guar} \\
\text{Regret guarantee:} \quad
\passk(\Variator) &\ge \passk(\Repeater) - (\tfrac{w}{4})^k. \label{eq:regret-guar}
\end{align}
\end{theorem}

\begin{figure}
    \centering
    \begin{subfigure}{0.34\textwidth}
        \centering
        \includegraphics[width=\textwidth]{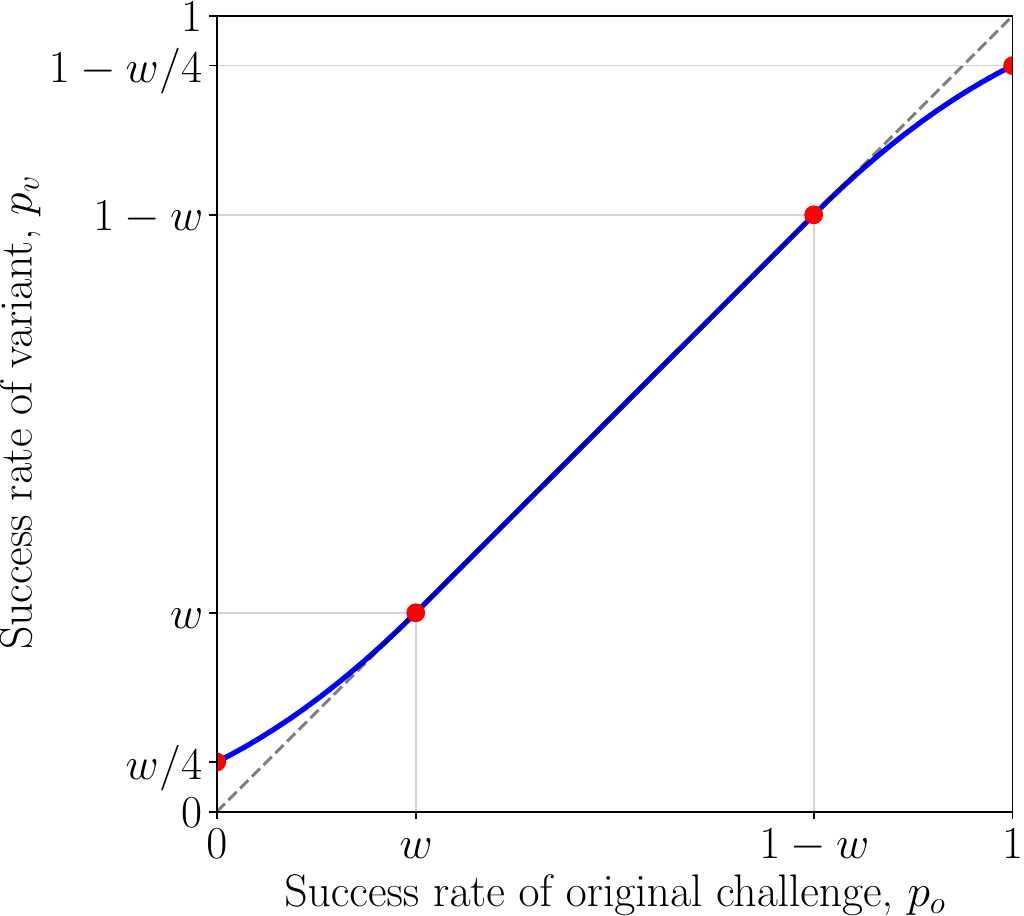}
        \caption{Expected variant success rate as a function of the original challenge success rate.}
        \label{fig:pv-po}
    \end{subfigure}
    \hspace{0.5em}
    \begin{subfigure}{0.305\textwidth}
        \centering
        \includegraphics[width=\textwidth]{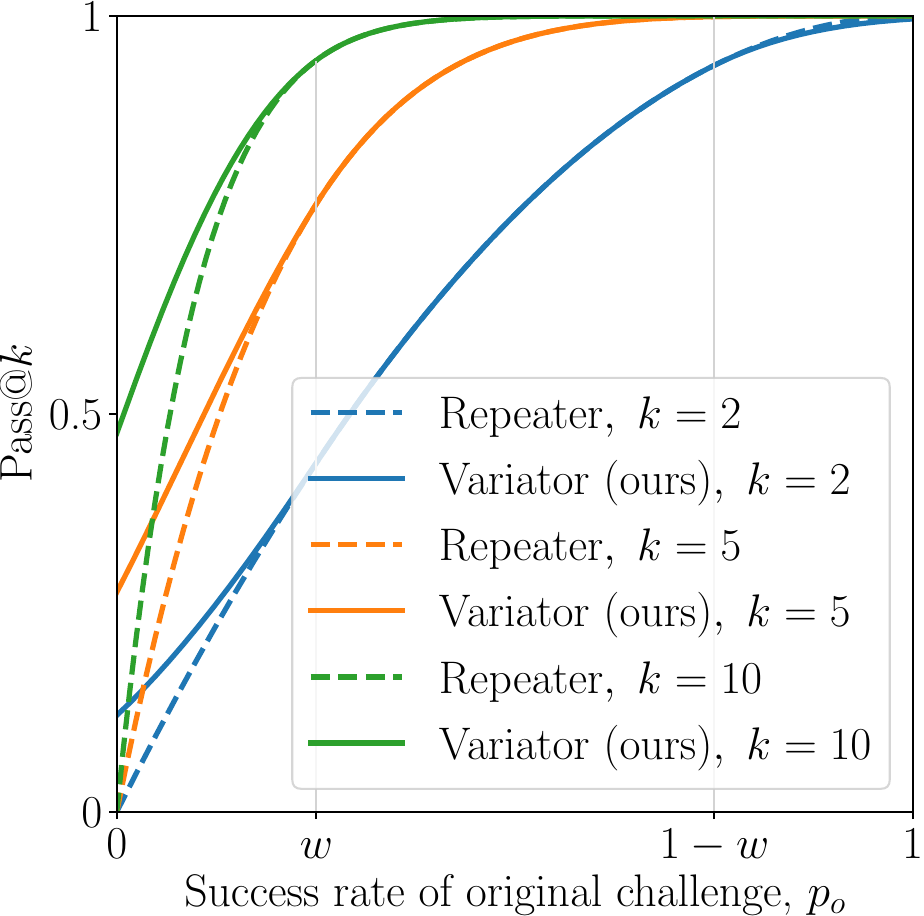}
        \caption{$\passk$ success rates using the Repeater agent and the Variator agent (ours).}
        \label{fig:pass_at_k}
    \end{subfigure}
    \hspace{0.5em}
    \begin{subfigure}{0.295\textwidth}
        \centering
        \includegraphics[width=\textwidth]{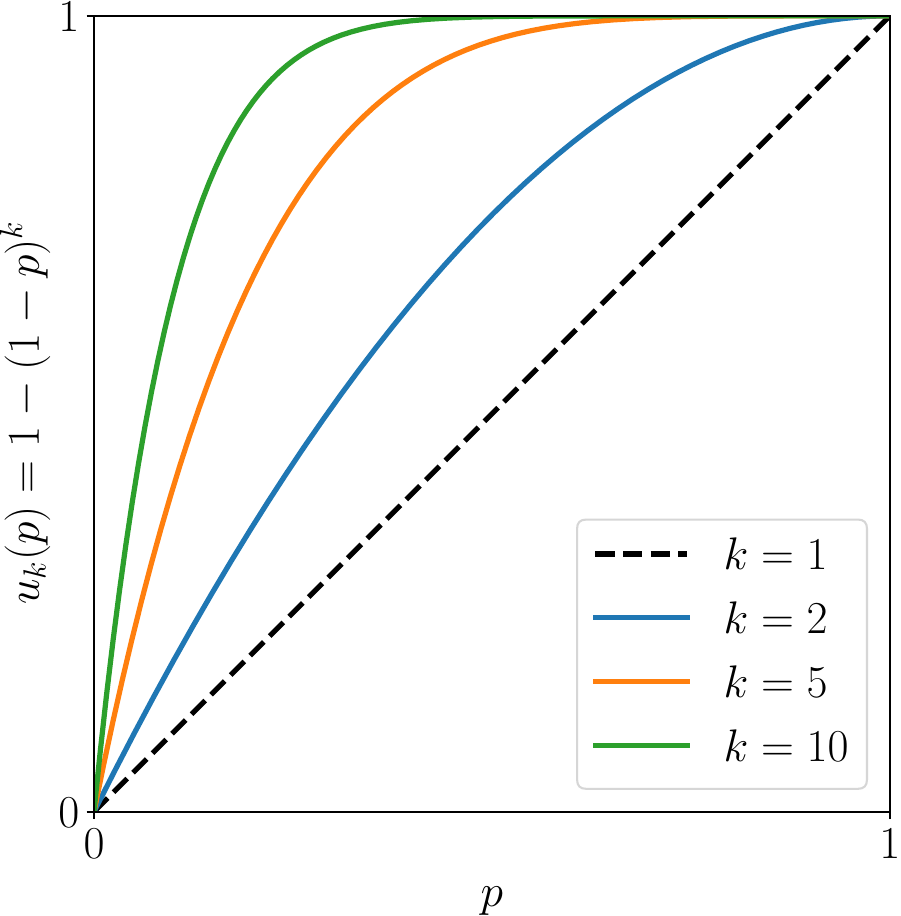}
        \caption{The function $u_k(p)$ pushes values $0<p<1$ towards $1$ when $k>1$.}
        \label{fig:u_k}
    \end{subfigure}
    \caption{Success rates of a challenge and its variants.}
    \label{fig:theory}
\end{figure}

Theorem~\ref{thm:guar} ensures that, for sufficiently high $k$, the performance of Variator approaches $1$ (eq. \eqref{eq:perf-guar}), and that it is always nearly as high as Repeater (eq. \eqref{eq:regret-guar}). These guarantees hold regardless of the value of $p_o$, and indeed are maintained even when $p_o = 0$, in which case Repeater's success rate is $0$. Moreover, the convergence is exponential with $k$, and so is quite rapid (Figure~\ref{fig:pass_at_k}): for example, with $w=0.2$, it is guaranteed that $\textrm{Pass@}10(\Variator) \ge 40\%$ and $\textrm{Pass@}10(\Variator) \ge \textrm{Pass@}10(\Repeater) - 10^{-13}$.

As this example demonstrates, Theorem~\ref{thm:guar} implies that if a benchmark contains challenges which are more or less evenly distributed over the entire range of success probabilities, then Variator's improvement on hard challenges will far outweigh its deterioration on easy challenges, resulting in an overall increase in $\passk$ when averaged over all challenges.

An outline of the proof is as follows (see full proof in Appendix~\ref{app:proof}). We begin by analyzing 
\begin{equation} \label{eq:p_v}
p_v \triangleq \mathbb{E}\{P_v\}
\end{equation}
which is the mean success rate of a random variant of $C$ (Figure~\ref{fig:pv-po}). The value of $p_v$ is monotonically increasing with $p_o$, and consists of three zones: For ``hard'' challenges, for which $p_o < w$, clipping at $0$ occurs, and we have $p_v > p_o$. Conversely, for ``easy'' challenges ($p_o > 1-w$), clipping at $1$ occurs and consequently $p_v < p_o$. Finally, for intermediate regions ($w \le p_o \le 1-w$), no clipping occurs and symmetry dictates that $p_v = p_o$.

Solving a single variant of $C$ is therefore not necessarily better than solving the original challenge: assuming we encounter both easy and hard challenges, some will perform better while others will be worse. The advantage of the Variator agent appears when considering the $\passk$ metric, for which it can be shown (see Appendix~\ref{app:proof}) that
\begin{align}
    \passk(\Repeater) &= 1 - (1-p_o)^k, \label{eq:passk-repeater} \\
    \passk(\Variator) &= 1 - (1-p_v)^k. \label{eq:passk-variator}
\end{align}
For $k>1$, the function $u_k(p) \triangleq 1-(1-p)^k$ pushes values $0<p<1$ towards $1$, exponentially in $k$ (see Figure~\ref{fig:u_k}). Thus, for hard challenges (low $p_o$), we have $p_v>p_o$ and this improvement is greatly emphasized by $u_k(\cdot)$. On the other hand, for easy challenges (high $p_o$), $p_v<p_o$, but this deterioration is negligible, because both $p_v$ and $p_o$ are pushed strongly towards $1$ by $u_k(\cdot)$ (see Figure~\ref{fig:pass_at_k}).

\subsection{Empirical results}
\label{sec:empirical}

In Section~\ref{sec:theoretical}, we showed analytically that, under simplified conditions, Variator's $\passk$ performance is never much worse than the baseline Repeater (Figure~\ref{fig:pass_at_k}). We now demonstrate that these results also hold empirically on a standard coding benchmark. We use the models described in Appendix~\ref{app:compute} and the APPS dataset described in Section~\ref{sec:challenges}. As this dataset is very large, to reduce computational costs we focus on a subset of the benchmark. Specifically, we restrict our analysis to all competition-level problems from the test set which include 60 or more test cases, resulting in a set of 60 challenges. 

To estimate the $\passk$ performance of each agent, we perform the following steps. We first generate 150 candidate solutions for the original challenge, from which the original success rate $p_o$ is computed. We then generate 25 variants using the technique described in Section~\ref{sec:generating}, and produce 6 candidate solutions for each variant (for a total of 150 candidates). The average success rate among these 150 candidates is our estimate for the mean variant success rate $p_v$ of eq.~\eqref{eq:p_v}. Finally, we compute $\passk$ for both agents from $p_o$ and $p_v$ using eqs.~\eqref{eq:passk-repeater} and \eqref{eq:passk-variator}.

\begin{figure}
    {\setlength{\tabcolsep}{1em}
    \centering
    \begin{tabular}[t]{p{0.42\textwidth}p{0.42\textwidth}}
    \begin{subfigure}[t]{0.42\textwidth}
        \centering
        \includegraphics[width=\textwidth]{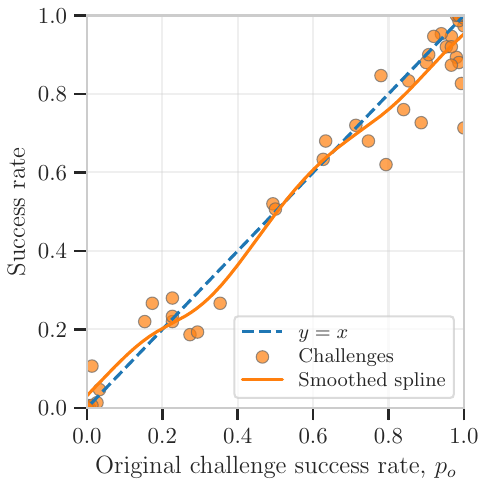}
        \caption{OpenAI o3-mini, public dataset}
        \label{fig:empirical-o3mini-public}
    \end{subfigure}
    &
    \begin{subfigure}[t]{0.42\textwidth}
        \centering
        \includegraphics[width=\textwidth]{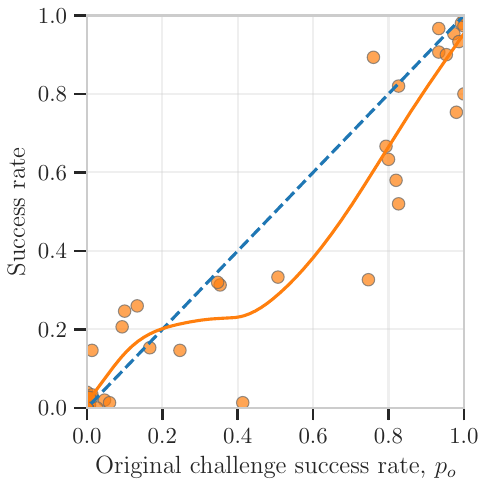}
        \caption{Claude 3.7 Sonnet (extended thinking),\\ public dataset}
        \label{fig:empirical-sonnet-public}
    \end{subfigure}
    \\[2em]
    \begin{subfigure}[t]{0.42\textwidth}
        \centering
        \includegraphics[width=\textwidth]{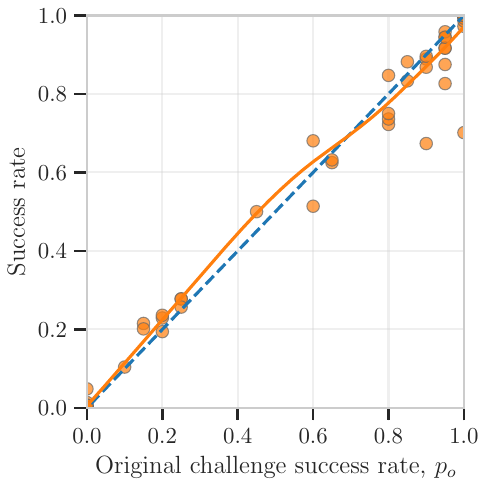}
        \caption{OpenAI o3-mini, private dataset}
        \label{fig:empirical-o3mini-private}
    \end{subfigure}
    &
    \begin{subfigure}[t]{0.42\textwidth}
        \centering
        \includegraphics[width=\textwidth]{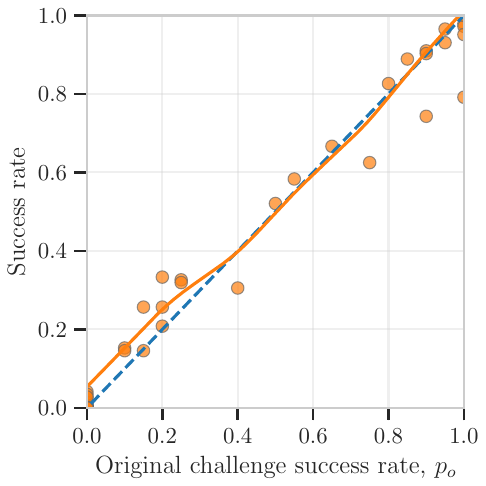}
        \caption{Claude 3.7 Sonnet (extended thinking),\\ private dataset}
        \label{fig:empirical-sonnet-private}
    \end{subfigure}
    \end{tabular}
    
    \caption{Average variant success rate $p_v$ as a function of the original challenge success rate $p_o$ for different models. Orange circles represent individual challenges, and the solid orange line is a smoothed spline interpolation thereof. Model memorization is one likely reason for reduced performance on the public dataset. This effect is no longer in play in the private dataset.}
    \label{fig:empirical}
    }
\end{figure}

The circles in Figs.~\ref{fig:empirical-o3mini-public} and \ref{fig:empirical-sonnet-public} illustrate $p_v$ as a function of $p_o$ for each challenge, while the solid orange line shows a smoothed spline interpolation of these individual points. As can be seen in these plots, $p_v$ tends to be slightly higher than $p_o$ for hard challenges (bottom-left corner), but underperforms elsewhere. Thus, on average, a single variant is likely to perform worse than the original challenge. This is evident in the $k=1$ column of Table~\ref{tab:modelpasskperformance}, since for $k=1$, Repeater submits a single candidate for the original challenge while Variator submits one candidate for a single variant.

Some practical considerations may explain this performance reduction. First, on public benchmarks, models are known to memorize challenges, even from the test set~\cite{li2024perteval}. As a result, the original challenge will have higher success rate than its variants, a somewhat artificial result of the fact that the model has previously been exposed only to the original challenge. Second, the occasional generation of non-equivalent variants will cause the LLM to sometimes solve a different problem than the original challenge, leading to incorrect submissions from Variator; note, though, that only around 6\% of generated variants are non-equivalent (Section~\ref{sec:generating}), so this effect is likely limited.

However, for larger $k$ values, these considerations become smaller than the gains provided by the $\passk$ effect of Theorem~\ref{thm:guar}. Consequently, as seen in Table~\ref{tab:modelpasskperformance}'s public dataset rows, Variator outperforms Repeater for $k \ge 5$ (for o3-mini) and $k \ge 10$ (for Sonnet 3.7). Evidently, Variator's performance improvement is strong enough that it overcomes the headwinds due to memorization, and exceeds the mean $\passk$ performance of Repeater.

\begin{table}[tbp]
{\setlength{\tabcolsep}{4pt}
\centering
\caption{Mean pass@k performance by model and agent, for the public and private datasets.}
\begin{tabular}{lllccccc}
\\
\toprule
\textbf{Dataset} & \textbf{Model} & \textbf{Agent} & $k=1$ & $k=5$ & $k=10$ & $k=15$ & $k=20$ \\
\midrule
\multirow{4}{*}{Public}
& \multirow{2}{*}{\shortstack[l]{Claude 3.7 Sonnet\\ {\small (extended thinking)}}}
  & Repeater & \textbf{29.80\%} & \textbf{40.38\%} & 44.05\% & 46.08\% & 47.46\% \\
&  & Variator (ours) & 26.53\% & 40.25\% & \textbf{44.62\%} & \textbf{46.67\%} & \textbf{48.00\%} \\
\cmidrule{2-8}
& \multirow{2}{*}{OpenAI o3-mini} 
  & Repeater & \textbf{57.14\%} & 70.36\% & 73.64\% & 74.96\% & 75.71\% \\
&   & Variator (ours) & 55.30\% & \textbf{70.89\%} & \textbf{74.50\%} & \textbf{75.87\%} & \textbf{76.57\%} \\
\midrule
\multirow{4}{*}{Private}
& \multirow{2}{*}{\shortstack[l]{Claude 3.7 Sonnet\\ {\small (extended thinking)}}}
  & Repeater & 26.17\% & 37.23\% & 40.78\% & 42.13\% & 42.73\% \\
&  & Variator (ours) & \textbf{26.50}\% & \textbf{40.30}\% & \textbf{44.72}\% & \textbf{46.81}\% & \textbf{48.16}\% \\
\cmidrule{2-8}
& \multirow{2}{*}{OpenAI o3-mini} 
  & Repeater & \textbf{56.42}\% & 69.57\% & 72.94\% & 74.12\% & 74.59\% \\
&   & Variator (ours) & 55.25\% & \textbf{70.97}\% & \textbf{74.57}\% & \textbf{75.93}\% & \textbf{76.63}\% \\
\bottomrule
\end{tabular}
\label{tab:modelpasskperformance}
}
\end{table}

The memorization effect is a result of leakage of the test set into models' training data, and is not expected to occur on novel, non-public datasets. Thus, we hypothesize that on such datasets, Variator's improvement will be even larger. To test this hypothesis, we repeat the above experiment using one of the variant challenges as the ``original'' challenge. Specifically, for each variant of each of the challenges described above, we compute the success rate by generating and testing 20 candidate solutions. We then select the variant with the median success rate to replace the original. This approach ensures that neither agent gains a systematic advantage. The original success rate $p_o$ is recalculated using 20 candidate solutions of the median variant. The mean variant success rate $p_v$ is recalculated using the remaining 24 variants. 

We refer to this set of challenges as a ``private dataset'' since none of the challenges could have appeared in a model's training data. The results are plotted in Figures~\ref{fig:empirical-o3mini-private} and \ref{fig:empirical-sonnet-private} and show striking similarity to the theoretical curve of Figure~\ref{fig:pv-po}, indicating that memorization of the public challenges may indeed have affected performance of the original experiment. The results of the private dataset experiment are shown in Table~\ref{tab:modelpasskperformance}. For $k=1$, as expected, neither agent is consistently better. For larger $k$ values, our method significantly outperforms the baseline; for example, using Claude 3.7, Variator improves on Repeater by 3--5 percentage points. These results further demonstrate the inconsistency effect's widespread presence in reasoning models, as well as the ability to utilize this effect to improve agent performance.

\section{Discussion and future work}
\label{sec:conclusion}

In this paper, we have demonstrated that inconsistency of LLMs across domains and models need not always be a drawback. Rather, we show that inconsistency can be leveraged to enhance model performance under the $\passk$ metric. This is achieved through the use of our novel Variator agent, whose advantages we demonstrate both theoretically and empirically. The $\passk$ metric, while applied in this work specifically for coding challenges, extends to many other practical scenarios where solutions can be easily verified, a common situation in many domains. Our variant generation method, extending paraphrasing to backstory alterations, is suitable for open-ended questions, making it applicable to numerous types of tasks, such as coding, cybersecurity, and mathematics.

While our approach shows promising results, we acknowledge certain limitations. First, our variation generation method may not benefit tasks with short descriptions, as the generated variants might be too similar. Second, in this paper, we focus on performance and do not prioritize computational efficiency. Yet, it is worth mentioning that the Variator agent requires an additional step of variant generation, increasing cost in both tokens and time. One way to mitigate this limitation might be to generate variants using a smaller and less costly model. Third, we demonstrate the inconsistency effect in two reasoning models, suggesting that this effect may persist in future models as well. Still, this assumption should be re-examined as new models are released.

Future extensions to our work may include an automatic verification step for the variants generated by the Variator to test equivalence, potentially further increasing the success rate of the Variator agent. An interesting research direction beyond this work’s scope is the use of our variant generation method to enrich models’ training sets, possibly leading to better generalization abilities. 

Lastly, further research into the mechanism behind the inconsistency effect is needed both to enhance model robustness and to optimize utilization of this phenomenon. The LLM response generation process can be conceptualized as a traversal through a tree of potential response pathways, some of which result in correct solutions while others terminate in erroneous outcomes. We hypothesize that some critical decision points in this tree are located near the beginning of the path, which may partly explain the impact of input wording on performance. Preliminary evidence supporting this hypothesis is presented in Appendix~\ref{app:inducing}, which we hope will assist subsequent research endeavors in this domain.

\section*{Acknowledgments}

We are extremely grateful to our panel of expert reviewers, consisting of Idan Geraffi, Maya Tamir, Tomer Fichman, and Yagil Markusfeld, who painstakingly reviewed each of the LLM-generated variants used in Section~\ref{sec:inconsistency-effect} in order to measure the ability of LLMs to generate equivalent variants. We also wish to thank Gil Gekker, Yoni Rozenshein, and Omer Antverg for their thoughtful feedback on earlier versions of this manuscript, which helped shape the final paper. Finally, we extend our gratitude to Mor Geva Pipek for helpful discussions and suggestions.

\newpage
\bibliography{references}

\newpage
\appendix
\section*{Appendices}

\section{Model settings and resources}
\label{app:compute}
To investigate the phenomenon of inconsistency in frontier models with reasoning abilities, we evaluated two state-of-the-art language models: Anthropic Claude 3.7 Sonnet (in extended thinking mode)~\cite{anthropic2025sonnet37}, and OpenAI o3-mini~\cite{openai2025o3mini}. In all of our experiments, the temperature of both models was set to 1, to improve performance and allow variability in the generation of responses.

The experiments were performed on a standard MacBook Pro with Apple M3 Pro chip and 18GB RAM, and on multiple EC2 servers by Amazon Web Services, configured as Ubuntu, ``t2.large'' with 2 vCPI and 8GB RAM.  We utilized cloud-based API services for accessing large language models:
\begin{itemize}
    \item AWS Bedrock interface~\cite{aws_bedrock} was used to access Anthropic's Claude 3.7 Sonnet. We used a limit of 4,000 thinking tokens and 10,000 total tokens.
    \item OpenAI's API interface~\cite{openai_api} was used for the coding experiemnts with OpenAI's o3-mini. Medium reasoning effort was used.
    \item Microsoft Azure's OpenAI Service~\cite{azure_openai} was used to generate results for the cybersecurity experiments of OpenAI's o3-mini. 
\end{itemize}

Coding tasks were run in parallel (up to 50 processes). Each coding challenge is solved within 15 minutes.
Cybersecurity challenges were also run in parallel (up to 15 challenges in parallel). The worst case solution time for CTF challenges, when the LLM exhausts all of its iterations, may take up to 50 minutes.

\section{Cybersecurity CTF challenges}
\label{app:cyber_challenges}

Due to wide differences in cyber capabilities between the two tested models (OpenAI's o3-mini \cite{openai2025o3mini} and Claude 3.7 Sonnet in extended thinking mode \cite{anthropic2025sonnet37}), we were unable to find a single challenge which was neither far below nor far above the capabilities of either model. Instead, to demonstrate the inconsistency effect, we supply two different cybersecurity CTF challenges, one for each model.

Both challenges are modeled around a classic cybersecurity CTF format, in which the LLM is allowed to interact with a machine running a vulnerable server application in a sandboxed environment. We limit the number of interactions to 20 and the time span of each interaction to two minutes. Successful task completion requires exploiting the vulnerability to retrieve a flag from the server. In both cases, a ``white-box'' scenario is implemented, wherein the LLM is allowed to examine the source code of the server application, to make it easier to find the vulnerability.

For Claude 3.7 Sonnet, we provide a challenge based on a common Remote Code Execution vulnerability occurring when deserializing untrusted data using Python's \verb|pickle| library~\cite{pickle}. The solution involves sending a ``malicious'' pickled file to the server, which, when unpacked, runs arbitrary code on the server in order to send the required flag back to the attacker.

For o3-mini, we provide a challenge based on extracting a flag from a poorly-designed keystore server. Specifically, the target flag is stored on the server and can be retrieved by supplying a four-digit password. There is no limit on the number of retrieval attempts, but there is a short delay after each incorrect request. Consequently, simply trying all 4-digit passwords serially would not succeed in the allotted time. The solution involves sending multiple parallel requests to the server, thus reducing the time until the correct password is found.

\section{Variant generation}
\label{app:variant_generation}
This appendix provides the complete text of the prompts we used to generate variants, as described in Section~\ref{sec:generating}.

To ensure diversity across our generated task variants, we implemented a sequential generation protocol requiring each variant to include a distinctive title that promotes contextual diversity. In the case of coding problems, we systematically compare the title of each new generated variant against the set of previously generated variants. Upon detection of an exact title match, the redundant variant is discarded. For the cybersecurity CTF variants, we employed an iterative prompt enhancement approach wherein the prompt is updated after each generation to include all titles of previously generated variants, and the model is explicitly instructed to avoid repeating the same theme.

\paragraph{Coding challenges variants.} For coding challenges, we used the following prompt to generate equivalent variants.

\begin{quote}
    \itshape The following is a coding competition exercise. I would like you to rewrite it in such a way that the solution to the exercise remains unchanged, but so that the new version seems completely different.
    Be original! Make up a new and completely different backstory, not the first thing that comes to mind.
    You may change the order in which the problem is described, the backstory, the wording, the mathematical notation, the terminology, the names of variables, and the formatting.
    However, the challenge must maintain the same input and output formats, so that a solution to the original challenge will also solve the new challenge.
    
    Please write the text of the new exercise within \verb|<challenge>| \dots \verb|</challenge>| tags.
    
    In addition, please give the new exercise a new and supply it within \verb|<title>| \dots \verb|</title>| tags.
    
    The original exercise is as follows:
    \normalfont{\texttt{\{the original coding challenge\}}}

\itshape
    Please provide your variation now, including both the <challenge> and <title> tags, without stopping for confirmation.
\end{quote}

Two examples of equivalent variants generated using this prompt, and used in the experiment described in Section~\ref{sec:inconsistency-effect}, are provided as an example in Table~\ref{table:coding-variant}.

\begin{table}[tbp]
    {\setlength{\tabcolsep}{8pt}
    \centering
    \caption{Two LLM-generated, equivalent variants of a coding challenge from the APPS dataset \cite{hendrycks21apps}. (Both versions are followed by specifications of the input and output format, as well as examples of inputs and corresponding outputs, which are not shown here.) The two variants are equivalent because a solution to either variant also solves the other, and they are equally difficult to solve according to human experts. Yet o3-mini has substantially higher success rates when solving the right challenge. The original challenge has a success rate of 63\% (CI: 55-71\%).}
    \begin{tabular}{p{0.46\textwidth} p{0.46\textwidth}}
    \\
    \toprule
    Variant: \textbf{Mosaic Mastery} & Variant: \textbf{Galactic Corridor Chromatics} \\
    \midrule
{\small In the storied realm of Divisorica, a renowned mosaic craftsman named Elyon has been commissioned to design a visually striking stone pathway for the royal palace.}
&
    {\small In the far reaches of space, Commander Zorin is reconfiguring the corridors of his interstellar vessel.}
\\
{\small The pathway is built from $n$ consecutive stones, sequentially numbered from $1$ to $n$.} &
    {\small The main corridor is neatly sectioned into $n$ consecutive segments, marked $1$ through $n$.}
\\
{\small Elyon wishes to adorn each stone with a splash of color.} &
    {\small Zorin has a collection of paints, each bearing a unique hue, and he plans to decorate each segment with one of these colors.}
\\
{\small However, the royal design guidelines impose a peculiar rule: if the gap between any two stones---say stones numbered $p$ and $q$---is such that the difference $d = |p-q|$ (with $d > 1$) exactly divides the total number of stones $n$ (i.e., $n \bmod d = 0$), then those two stones must be painted in the same hue.}
&
    {\small However, the Commander has an unconventional aesthetic rule. He insists that for any two segments at positions $a$ and $b$ (with $a \ne b$) that are separated by a distance $d$ (i.e.\ $|a - b| = d$) where $d > 1$ and $d$ evenly divides $n$ (meaning $n \bmod d = 0$), the segments must share the identical color. This rule is set to maintain a mysterious symmetry in the vessel’s design.}
\\
{\small Elyon’s goal is to maximize the variety in his palette, while still following these design mandates. He wonders: What is the greatest number of distinct colors he may use in painting the pathway under the rules specified?}
&
    {\small Your task is to determine the greatest number of distinct colors Zorin can employ while ensuring the corridor remains visually “balanced” according to his rule.}
\\
    \midrule
    o3-mini: 28\% success (CI: 22--34\%) & 
    o3-mini: 80\% success (CI: 74--85\%) \\
    \bottomrule
    \end{tabular}
    \vspace{3pt}
    \label{table:coding-variant}}
\end{table}

\paragraph{Cybersecurity CTF challenges variants.} For cybersecurity CTF challenges, we used the following prompts to generate equivalent variants. To create variants of the Pickle vulnerability challenge for Claude 3.7 Sonnet, we used the following prompt: 
\begin{quote}
    \itshape I am writing CTF cyber challenges
    
    I will provide you with the content of two files :
    
    app.py - the application that runs the CTF challenges
    
    solution.py - a valid solution that solves this cyber CTF challenge
    
    I want you to recreate app.y - so that I will have a completely different context - change everything you need that this application will have another theme

    Please provide a new code, that essentially does the same but still I want the newly created code to be not easily recognizable as another version of the original CTF challenge. As an explicit instruction : completely redesign the code architecture using different design patterns.

    But, the solution code must solve the CTF you write as well - so pay attention to function recvuntil and the strings it is expecting - the strings must match.
    
    Also, please provide a title for the new theme you assign, use up to four words, in CamelCase.

    Also, please provide a short, up to five words description of the architecture you used for the application code.

    Avoid using the themes : \normalfont{\texttt{\{previously used theme titles\}}}\itshape
    
    Avoid using code structures such as : \normalfont{\texttt{\{previously used design patterns\}}}\itshape

    app.py: \verb|{app.py content}|

    solution.py: \verb|{solution.py content}|
\end{quote}

To create variants of the ``online enumeration'' challenge for o3-mini, we used the following prompt:

\begin{quote}
\itshape 
\# CTF Challenge Transformation

\#\# Task Overview

Create a transformed version of a Capture The Flag (CTF) cyber security challenge, which appears to participants to be an entirely different challenge, but is in fact solved using the exact same code which solves the original challenge.

\#\# Input Files

You will receive two files:

1. `app.py`: The server-side application that powers the CTF challenge

2. `solution.py`: A working exploit script that successfully solves the challenge

\#\# Detailed Requirements

\#\#\# Theme Transformation

* Develop a completely new thematic context

* Replace all theme-specific terminology, visuals, and narrative elements

* Create appropriate themed feedback messages and user interface elements

\#\#\# Architectural Redesign

* Implement a significantly different code architecture and design patterns

* Reorganize code structure, class hierarchies, and component interactions

* Use different programming paradigms where appropriate (functional, OOP, etc.)

* Restructure the code organization so that a side-by-side comparison would not reveal similarities

\#\#\# Functional Preservation

* Maintain identical validation logic for the core exploitation mechanism

* Preserve the exact behavior of credential verification

* Ensure all security level transitions occur at the same thresholds

* Retain the same flag retrieval mechanism

\#\#\# Solution Compatibility

* The original `solution.py` must work with your new implementation without any changes

* All network communication patterns must remain identical

* The challenge must respond to the same sequence of inputs in the same way

* Flag output format must remain consistent with the original

\#\#\# Obfuscation Requirements

* Avoid variable names, comments, or patterns that hint at the original challenge

* Do not include comments that suggest this is a transformed version of another challenge

* Replace all function and variable names with theme-appropriate alternatives

\#\# Deliverables

1. Complete code for the transformed `app.py`

2. A title for your new theme (CamelCase format, maximum 4 words)

3. Brief description of your architectural approach (maximum 5 words)

\#\# Constraints

* Avoid making only superficial thematic changes to the provided code - Use a completely different design pattern
* **Banned themes**: 
\normalfont{\texttt{\{previously used themes placeholder\}}}\itshape

* **Banned design patterns**: 

\normalfont{\texttt{\{previously used design patterns placeholder\} }}\itshape

app.py:
\verb|{app.py content}|

solution.py:
\verb|{solution.py content}|

Reminder - Avoid making only superficial thematic changes to the provided code - Use a completely different design pattern
\end{quote}

\subsection{Manual verification}
\label{app:manual_verification}
As mentioned in Section~\ref{sec:experimental_setup}, the generated variants used for the demonstrations in Section~\ref{sec:inconsistency-effect} have been manually validated by professional cybersecurity experts. In this appendix, we provide the instructions given to these experts. In addition to the instructions, the human experts were given a list of task variants, including a title for each task, where the original task was marked as such.

The following instructions were given to experts validating the variants of the coding challenges:
\begin{quote}
    Please review the following files. Each file represents a coding challenge. The challenges should be very similar. Please review all of the challenges.
We expect the challenges to have very similar difficulty levels, if you find a challenge which is easier/harder please mark it and explain why.
Please feel free to comment any insight in that regard.

\end{quote}
The following instructions were given to experts validating the variants of the coding challenges:
\begin{quote}
    Please review the following files. Each file represents a server that is used as an application for a CTF challenge.
The file solution.py solves all of the challenges listed here.
We expect the challenges to have very similar difficulty levels, if you find a challenge which is easier/harder please mark it and explain why.
\end{quote}

\section{Inconsistency effect: Experiment details}
\label{app:inconsistency-experiments}

In this appendix, we present the results of the experiments demonstrating the inconsistency effect on frontier reasoning models, as described in Section~\ref{sec:inconsistency-effect}. The effect was demonstrated on both coding tasks (APPS test-set tasks numbered 4638 and 3743 for Claude 3.7 and o3-mini respectively) and cybersecurity CTF challenges (Appendix~\ref{app:cyber_challenges}). Table \ref{tab:standalone_experiments} summarizes the experiment statistics, including: the number of variants generated; of those, the number of variants which were rejected by experts because they were deemed non-equivalent; the number of solutions generated for each variant; and the $p$-value of the distribution of success rates for those challenges, under the null hypothesis that the inconsistency effect does not hold (see Section~\ref{sec:inconsistency-effect}).

Notably, over 95\% of variants were approved as equivalent (see Table~\ref{tab:standalone_experiments}). Moreover, model success rates on excluded variants were near the average performance of equivalent variants, indicating that experts tended to be cautious and excluded even variants which were not meaningfully more difficult than the original.

\begin{table}[htbp]
{\setlength{\tabcolsep}{4.5pt}
\caption{Summary statistics for an experiment demonstrating the inconsistency effect on frontier reasoning models, for coding and cyber tasks. The $p$-value represents the likelihood of the observations under the null hypothesis that the inconsistency effect does not hold.}
\centering
\begin{tabular}{llp{1.5cm}p{1.8cm}p{1.8cm}p{1.5cm}}
\\
\toprule
\textbf{Model} & \textbf{Domain} & \textbf{Variants produced} & \textbf{Variants marked non-equiv.} & \textbf{Solutions per variant} & \textbf{$p$-value} \\
\midrule
\multirow{2}{*}{\shortstack[l]{Claude 3.7 Sonnet\\(extended thinking mode)}}
  & Cyber & 30 & 0 & 50 & $5\times10^{-5}$ \\
  & Coding & 25 & 1 & 50 & $3\times10^{-6}$ \\
\midrule
\multirow{2}{*}{\shortstack[l]{OpenAI o3-mini}}
  & Cyber & 30 & 0 & 50 & $8\times10^{-5}$ \\
  & Coding & 25 & 2 & 50 & $4\times10^{-4}$ \\
\midrule
\end{tabular}
\label{tab:standalone_experiments}
}
\end{table}

\subsection{Hint experiment}
\label{app:hint_experiment}
In this appendix, we present the prompts we used in order to generate coding solutions, and coding solutions for the hint-enhanced variants. The experiment was conducted on Claude 3.7 Sonnet (extended thinking mode) with task number 4638 from the APPS benchmark.

\paragraph{Basic prompt for solving coding challenges.}
\begin{quote}
    \itshape Can you write Python code to complete the following coding task?
    
    Do not provide explanations. Provide your Python code enclosed by \verb|<solution>| \dots \verb|</solution>| tags, without markdown.
    
    The code you provide will be executed directly, so be sure to include an \verb|`if __name__ == '__main__':`| block.
    
    The code you write must complete the task within 2 seconds and may use up to 256MB RAM.
    
    \normalfont{\texttt{\{a description of the coding challenge\}}
    
    \texttt{\{hint placeholder\}}}\itshape
    
    Remember, the code you write must complete the task within 2 seconds and may use up to 256MB RAM, so make sure it is efficient both in runtime and memory utilization.
\end{quote}

\paragraph{Hint placeholder content.}
\begin{quote}
    \itshape As a hint, a description of a possible solution is provided: \normalfont{\texttt{\{specific hint content\}}}
\end{quote}

\paragraph{Provided hints.}
\begin{quote}
(a) \itshape Use dynamic programming.
    
\normalfont{(b)} \itshape The solution implements Dijkstra's algorithm to find minimum-cost paths from a source node to all other nodes in a graph. What makes this problem interesting is that nodes can be connected directly to any other node (not just adjacent ones), with the cost calculated by summing intermediate segment costs. For each path between nodes, the algorithm compares two different cost calculation methods: a direct sum of segment weights versus an alternative that adds a fixed overhead cost plus a different set of segment weights. By using a priority queue to process nodes in order of increasing cumulative cost, the algorithm guarantees optimal solutions for all destination nodes.
\end{quote}

These hints were generated by a professional software developer based on the nature of the coding challenge under investigation. As a result, the hints improved the average success rate of variants from 60\% to 68\% with hint (a) and 87\% with hint (b), respectively.

\section{Proof of Theorem~\ref{thm:guar}}
\label{app:proof}

\begin{proof}
Consider first the value $\passk(\Repeater)$, which is the probability that at least one of the $k$ solutions generated by the LLM are correct. Since the probability of a correct single solution is $p_o$, we have $\passk(\Repeater) = 1 - (1 - p_o)^k$, thus proving \eqref{eq:passk-repeater}.

Now consider the Variator agent, which generates $k$ variants of $C$. Let us denote the success rate of the $j$th variant by $[p_o + W_j]_0^1$, where $W_1, \ldots, W_k$ are independent and identically distributed random variables which are uniformly distributed in the range $[-w, w]$. The probability that at least one of these variants will be solved correctly is then
$$
\E { 1 - \prod_{i=1}^k (1 - [p_o + W_j]_0^1) }
= 1 - \prod_{i=1}^k \E { (1 - [p_o + W_j]_0^1) }
$$
where we used the fact that the $W_j$'s are independent. Denoting 
\begin{equation} \label{eq:pv}
p_v \triangleq \E{[p_o + W]_0^1},   
\end{equation}
we obtain \eqref{eq:passk-variator}.

It follows that the metrics $\passk(\Repeater)$ and $\passk(\Variator)$ depend solely on the values $p_o$ and $p_v$, respectively. Now, observe that $\E{p_o + W} = p_o$, so that the only difference between $p_o$ and $p_v$ is due to clipping: specifically, clipping at $0$ increases $p_v$ relative to $p_o$, whereas clipping at $1$ decreases $p_v$ relative to $p_o$. This is illustrated in Figure~\ref{fig:pv-po}, which plots the actual dependence between $p_v$ and $p_o$ as derived below.

Let us therefore compute the value of $p_v$ as a function of $p_o$. This requires careful attention to the effect of clipping at $0$ and $1$, but is otherwise a straightforward exercise in probability theory. Consider first the case $w \le p_o \le 1-w$. In this case, we have $0 \le p_o + W \le 1$, so that clipping does not occur; hence, by \eqref{eq:pv}, we have $p_v = p_o$.

Next, define $Z \triangleq [p_o + W]_0^1$ and consider the case $0 \le p_o < w$. In this case,
$$
Z \sim \begin{cases}
    0,             & \text{w.p. } \frac{w - p_o}{2w}, \\
    U[0, w + p_o], & \text{w.p. } \frac{w + p_o}{2w}.
\end{cases}
$$
Therefore
$$
p_v = \E{Z} = 0 \cdot \frac{w - p_o}{2w} + \frac{w + p_o}{2} \cdot \frac{w + p_o}{2w}
= \frac{(w + p_o)^2}{4w}.
$$

Finally, and consider the case $1-w < p_o \le 1$. Using similar reasoning, in this case we have
$$
Z \sim \begin{cases}
    1,             & \text{w.p. } \frac{p_o + w - 1}{2w}, \\
    U[p_o - w, 1], & \text{w.p. } \frac{w+1-p_o}{2w}.
\end{cases}
$$
Therefore
$$
p_v = \E{Z} = 1 \cdot \frac{p_o + w - 1}{2w} + \frac{1+p_o-w}{2} \cdot \frac{w+1-p_o}{2w}
$$
which simplifies to
$$
p_v = 1 - \frac{(1 - p_o + w)^2}{4w}.
$$

Combining these results, we have
\begin{equation} \label{eq:pv-res}
p_v = 
\begin{cases}
    \frac{(w + p_o)^2}{4w}         &\text{for } 0 \le p_o < w, \\
    p_o                            &\text{for } w \le p_o \le 1-w, \\
    1 - \frac{(1 + w - p_o)^2}{4w} &\text{for } 1-w < p_o \le 1.
\end{cases}
\end{equation}
It can readily be verified that $p_v$ is continuous and monotonically increasing in $p_o$, and thus its minimum and maximum in the range $[0,1]$ are obtained at $p_o=0$ and $p_o=1$, respectively. This is illustrated in Figure~\ref{fig:pv-po}. It follows that
\begin{equation}
    p_v \ge p_v \big|_{p_o=0} = \frac{w}{4}.
\end{equation}
Substituting this into \eqref{eq:passk-variator} yields the required \eqref{eq:perf-guar}. Likewise, it is readily seen that the largest difference $p_o - p_v$ is obtained when $p_o=1$. From \eqref{eq:passk-repeater} and \eqref{eq:passk-variator}, it follows that this is also the point at which the difference $\passk(\Repeater) - \passk(\Variator)$ is maximal. At this point, we have $p_v=1-w/4$ and $p_o = 1$. From \eqref{eq:passk-variator} and \eqref{eq:passk-repeater}, this implies
$$
\passk(\Repeater) - \passk(\Variator) = 1 - \left( 1 - \left( \tfrac{w}{4} \right)^k \right) = \left( \tfrac{w}{4} \right)^k
$$
which yields the required \eqref{eq:regret-guar}.
\end{proof}

\section{Discussion: Inducing failures}
\label{app:inducing}

In this paper, we have treated the inconsistency effect as an empirical phenomenon. We demonstrated that models regularly exhibit inconsistency in multiple domains, and that it is possible to utilize this phenomenon to gain meaningful performance boosts. In this section, we take a step back and describe an experiment which hints at the causal process leading to the generation of inconsistent responses.

We find it useful to envision the LLM response generation process as one in which the LLM progresses through a tree of potential response pathways. Nodes in the tree correspond to decision points where the model may choose between feasible response paths. Some of these decisions are functionally unimportant, such as wording choice, while other decisions are strategic, such as selecting a design pattern for a function. Among the strategic decisions in particular, some pathways will lead to correct solutions, while others include logical errors or poor assessments and will end up failing to solve the challenge.

To investigate this process, we demonstrate that failure to solve a challenge can be systematically induced in otherwise successful variants by prefilling the response with portions of a \emph{correct} solution, which the model then completes \emph{incorrectly}. We demonstrate that such strings can be constructed systematically by taking a correct prefix of the output from failing solutions of other variants, and applying them to successful variants. 

\subsection{Experimental setup}
To make the setting easier to follow, in this section we restrict attention to highschool-level verbal math problems, as such problems naturally trace out a step-by-step solution, where a mistake at any point leads to an incorrect solution thenceforth. We specifically consider the two variants shown in Table~\ref{table:ships-caterpillars}. These variants are equivalent because the verbal problems translate into the same set of mathematical equations, and they thus have the same numerical solution and the same difficulty level.

\begin{table}[tbp]
    {\setlength{\tabcolsep}{9pt}
    \caption{Two equivalent math problems with varying success rates in Claude Sonnet 3.5 (model version 2024-10-22)~\cite{anthropic2024sonnet35oct}. In both cases, the LLM correctly writes out the (equivalent) mathematical equations required to solve the problem, but consistently finds it more difficult to correctly transfer terms in the equations arising in the Ships problem.}
    \centering
    \begin{tabular}{p{0.44\textwidth} p{0.44\textwidth}}
    \\
    \toprule
    \textbf{Ships} & \textbf{Caterpillars} \\
    \midrule
    A large cruise ship and a sailboat both depart from Pearl Harbor at the same time, heading to a remote Pacific island 480 miles away. Both vessels maintain constant speeds throughout their journey. The sailboat arrives at the island 2 hours after the cruise ship, having traveled 20 miles per hour slower than the ship. What was the cruise ship's speed? & 
    A caterpillar pupates after eating 480 leaves. Two caterpillars hatch on the same day and start eating leaves at a constant rate. Every day, the first caterpillar eats 20 leaves more than the second caterpillar. Consequently, the first caterpillar pupates two days before the second caterpillar. Determine the number of leaves the first caterpillar eats per day. \\
    \midrule
    Claude: 69\% success (CI: [63\%, 74\%]) & 
    Claude: 96\% success (CI: [93\%, 98\%]) \\
    \bottomrule
    \end{tabular}
    \label{table:ships-caterpillars}}
\end{table}

\begin{table}[tbp]
    {\setlength{\tabcolsep}{3pt}
    \caption{``Translating'' a failing solution from one variant to another. The Ships solution is correct up until the line marked in red, in which the term \texttt{480x} is transferred incorrectly. The translated version stops a few steps before the math error, but the LLM will consistently perform that same error in the otherwise successful Caterpillars variant.}
    \centering
    \begin{tabular}{m{0.48\textwidth} m{0.48\textwidth}}
    \\
    \toprule
    \textbf{Ships: Typical failing solution} & \textbf{Caterpillars: Translated solution prefix} \\
    \midrule
    \parbox{0.48\textwidth}{\texttt{\small{\\
        Let me solve this step by step. \\
        1.\ Let's define variables: \\
        \phantom{1.\ }* Let x = cruise ship's \\ \phantom{1.\ * }speed in mph \\
        \phantom{1.\ }* Then (x-20) = sailboat's \\ \phantom{1.\ * }speed in mph \\
        \phantom{1.\ }* Distance = 480 miles \\ \phantom{1.\ * }for both vessels \\
        ... \\
        3.\ Multiply both sides by x: \\
        \phantom{1.\ }* 480x = (x-20)(480 + 2x) \\
        \phantom{1.\ }* 480x = 480x - 9600 + 2x² - 40x \\
        \phantom{1.\ }* \textcolor{red}{0 = 2x² - 520x - 9600} \\ }}
        \small \emph{(Incorrect solution continues\ldots.)}}
    & \parbox{0.48\textwidth}{\vspace{-5pt}\texttt{\small{Let me solve this step by step. \\
        1.\ Let's define variables: \\
        \phantom{1.\ }* Let x = first caterpillar's \\ \phantom{1.\ * }eating rate in leaves per day \\
        \phantom{1.\ }* Then (x-20) = second caterpillar's \\ \phantom{1.\ * }eating rate in leaves per day \\
        \phantom{1.\ }* Total leaves = 480 \\ \phantom{1.\ * }for both caterpillars \\
        ... \\
        3.\ Multiply both sides by x: \\ }}
        \small \emph{(LLM to continue response here.)}}
    \\
    \bottomrule
    \end{tabular}
    \label{table:translate}}
\end{table}

It can be seen that the Caterpillars variant is solved nearly perfectly (96\% success rate), while the Ships variant fails far more often (69\% success rate). Failures in the Ships variant occur as a result of a mathematical error, often an incorrect transfer-of-terms, as in the example on the left side of Table~\ref{table:translate}. To induce a failure in the Caterpillars variant, we perform the following (manual) steps:
\begin{enumerate}
    \item Select a failing solution of the Ships variant, such as the one on the left side of Table~\ref{table:translate}.
    \item Identify the first point of failure in the solution, such as the first position in which a term is incorrectly transferred (in our case, the line marked red in Table~\ref{table:translate}).
    \item Take a prefix ending somewhat earlier than the first point of failure. In our experiment, we ended the prefix some 50 tokens prior to the failure. This prefix is thus a correct partial solution of the challenge.
    \item Translate the prefix to the setting of the Caterpillars variant. In particular, replace every occurrence of ``mph'' to ``leaves per day,'' every occurrence of ``speed'' with ``eating rate,'' and so on. The result is a valid solution prefix for the Caterpillars variant, and can be seen in the right side of Table~\ref{table:translate}.
    \item Run the LLM on the Caterpillars variant, but prefill the response with the solution prefix. Sample the results multiple times to obtain an average success rate.
\end{enumerate}

\subsection{Results and conclusions}
When running the Caterpillars variant with the prefilled solution prefix, the model obtains a correct solution in only 36\% of the cases (CI: [28\%, 45\%]), far lower than the original Caterpillars success rate (96\%) and even lower than the Ships success rate (69\%). The failures most often result from the exact transfer-of-terms mistake observed in the left side of Table~\ref{table:translate}.

This experiment implies that moving onto a failure pathway does not seem to be accompanied by a human-detectable ``glitch'' in the output. Once on a failure pathway, the model may continue with multiple valid steps in the solution, but will ultimately perform a false logical leap.

We interpret this result as follows. The complex challenges faced by LLMs can be solved in many correct ways, e.g., different designs can be selected to solve a coding challenge. These are different pathways, among which the choice occurs early in the token stream. Since all such pathways can in principle lead to correct solutions, the decision among them is arbitrary to some extent. As we observe above, equivalent variants sometimes tend to prefer different pathways. However, some of these pathways have a higher probability of generating an error later in the output stream. This results in a different success probability for different variants. Under this interpretation, the experiment described above has forced the model to choose a pathway leading to failure in a variant which would generally pick a different pathway.

More research would be required to confirm or reject this hypothesis. In particular, as yet we do not know what causes the model to choose a particular pathway for each variant, nor can we predict which pathway will be taken for a given variant. We hope that this experiment provides a small step towards further understanding of the underlying processes.

\end{document}